\documentclass[letterpaper]{article} 
\usepackage{aaai2026}  
\usepackage{times}  
\usepackage{helvet}  
\usepackage{courier}  
\usepackage[hyphens]{url}  
\usepackage{graphicx} 
\usepackage{natbib}  
\usepackage{caption} 
\usepackage{algorithm}
\usepackage{algorithmic}
\usepackage{amsmath}
\usepackage{soul, color, xcolor}
\usepackage{threeparttable}
\usepackage{multirow}

\usepackage{newfloat}
\usepackage{listings}
\DeclareCaptionStyle{ruled}{labelfont=normalfont,labelsep=colon,strut=off} 
\floatstyle{ruled}
\newfloat{listing}{tb}{lst}{}
\floatname{listing}{Listing}
\title{ParaRevSNN: A Parallel Reversible Spiking Neural Network for Efficient Training and Inference}
\author{
Changqing Xu,
Guoqing Sun,
Yi Liu,
Xinfang Liao,
Yintang Yang\\
}

\affiliations{School of Microelectronics, Xidian University, Xi’an, China\\
\{cqxu, gqsun, yiliu, xfliao, ytyang\}@xidian.edu.cn}

\usepackage{bibentry}

\begin{document}

\maketitle

\begin{abstract}
{Reversible Spiking Neural Networks (RevSNNs) enable memory-efficient training by reconstructing forward activations during backpropagation, but suffer from high latency due to strictly sequential computation. To overcome this limitation, we propose ParaRevSNN, a parallel reversible SNN architecture that decouples sequential dependencies between reversible blocks while preserving reversibility. This design enables inter-block parallelism, significantly accelerating training and inference while retaining the memory-saving benefits of reversibility. Experiments on CIFAR10, CIFAR100, CIFAR10-DVS, and DVS128 Gesture demonstrate that ParaRevSNN matches or exceeds the accuracy of standard RevSNNs, while reducing training time by up to 35.2\% and inference time to 18.15\%, making it well-suited for deployment in resource-constrained scenarios.}


\end{abstract}

%

\section{Introduction}

Spiking Neural Networks (SNNs), inspired by the brain, represent the third generation of neural networks, offering high biological plausibility and strong robustness\cite{maass1997networks}. Due to their event-driven and spike-based computation paradigm, SNNs exhibit significant energy efficiency advantages when deployed on edge devices\cite{akopyan2015truenorth,davies2018loihi}. 

However, training SNNs remains a major challenge\cite{meng2022training,yang2023effective}. Since information in SNNs is transmitted through non-differentiable spike sequences\cite{wu2018spatio}, current supervised training methods often rely on surrogate gradient techniques to approximate gradients\cite{xu2023ultra,zhang2019spike}, enabling end-to-end optimization. While such methods outperform ANN-to-SNN conversion\cite{kim2022privatesnn,bojkovic2025temporal,aydin2024hybrid} in terms of latency and accuracy, they are still difficult to deploy on edge devices. This is primarily due to the use of backpropagation through time (BPTT) to capture spatiotemporal dynamics\cite{xiao2022online,meng2023towards}, which leads to substantial computational latency and memory\cite{zhang2025memory} consumption—posing serious obstacles for real-time and resource-constrained applications\cite{zhang2024memory}.

To address the challenges above, reversible computing techniques were first introduced in Artificial Neural Networks (ANNs), initially in the context of unsupervised learning\cite{dinh2014nice}. The key idea is to design reversible blocks that allow the reconstruction of forward activations during backpropagation, thereby significantly reducing memory usage\cite{gomez2017reversible,sander2021momentum,mangalam2022reversible,brugger2019partially}. This approach was later extended to the domain of Spiking Neural Networks\cite{zhang2024memory,hu2024high}, where it effectively mitigated the high memory demands associated with Backpropagation Through Time. However, existing reversible SNN architectures are typically constructed using sequentially stacked reversible blocks, which are inherently challenging to align with parallel hardware execution. As a result, the incorporation of reversible mechanisms often leads to increased training and inference latency, limiting their practicality for deployment on efficiency-critical hardware platforms.

To address the intrinsic serialization bottleneck in conventional reversible spiking neural networks, we propose ParaRevSNN (Parallel Reversible Spiking Neural Network), a novel architecture that enables intra-block parallelism while preserving reversibility. By redesigning the data dependency within reversible modules, ParaRevSNN breaks the tightly coupled forward and backward computation pattern, allowing critical paths to be computed in parallel. This design not only retains the memory efficiency benefits of reversible computation but also significantly improves computational throughput during both inference and training. Extensive experiments on multiple benchmark datasets demonstrate that ParaRevSNN consistently achieves high accuracy while considerably reducing memory consumption. Notably, the proposed method is validated across various datasets, demonstrating its versatility and strong generalization ability in diverse spiking learning tasks. Furthermore, ParaRevSNN exhibits promising performance in resource-constrained scenarios, making it well-suited for deployment on edge devices.
Our main contributions are summarized as follows:

\begin{itemize}\item We propose a novel ParaRevSNN-ResNet architecture that enables inter-block parallelism in reversible spiking neural networks, mitigating the inherent sequential bottlenecks during forward and backward computations.
\item Extensive experiments are conducted on various benchmark datasets, demonstrating the superiority of our method in terms of memory efficiency and training speed.
\item We further investigate the impact of network depth on performance, showing that the proposed architecture exhibits greater scalability and efficiency in deeper networks compared to conventional reversible SNNs.\end{itemize}


\section{Related work}
\subsection{Spiking Neuron Model}

This work focuses on two widely used spiking neuron models: the Integrate-and-Fire (IF) neuron and the Leaky Integrate-and-Fire (LIF) neuron. These models describe spiking dynamics over discrete time steps and serve as the foundation for our experimental framework.

\paragraph{IF Neuron.} 
The Integrate-and-Fire (IF) neuron updates its membrane potential \( V[t] \) based on the accumulated input current \( X[t] \). A spike is emitted when the membrane potential exceeds a threshold \( \theta \), and the membrane is reset:
\begin{align}
V[t] &= V[t-1] + X[t], \label{eq:if_voltage} \\
S[t] &= H(V[t] - \theta), \label{eq:if_spike} \\
V[t] &= V[t] \cdot (1 - S[t]), \label{eq:if_reset}
\end{align}
where \( H(\cdot) \) is the Heaviside step function, and \( S[t] \in \{0,1\} \) indicates the presence of a spike.

\paragraph{LIF Neuron.}
The Leaky Integrate-and-Fire (LIF) neuron introduces a decay term to simulate the membrane leak over time. The update rule is:
\begin{align}
V[t] &= \alpha V[t-1] + X[t], \label{eq:lif_voltage}
\end{align}
where \( \alpha \in (0,1) \) is the decay factor. Spike generation and reset follow:
\begin{align}
S[t] &= H(V[t] - \theta), \label{eq:lif_spike} \\
V[t] &= V[t] \cdot (1 - S[t]). \label{eq:lif_reset}
\end{align}

In the Parametric LIF neuron, the decay factor \( \tau \) is predefined rather than learned during training. This fixed temporal decay governs the membrane potential dynamics, providing stable temporal behavior across tasks without introducing additional trainable parameters.

\subsection{Efficient Training Methods}
To address the non-differentiability of SNNs, a widely adopted strategy is to replace the spike function with a surrogate gradient, thereby enabling backpropagation in non-continuous neural models\cite{lee2016training}. Although this approach can achieve competitive accuracy with relatively short temporal sequences, it incurs substantial memory and latency overhead due to the need to store intermediate states across all time steps during training\cite{deng2020rethinking}.
To mitigate this issue, various optimization strategies have been proposed from different perspectives. Some works focus on reducing the number of timesteps to decrease computational complexity. For example, dynamic timestep adjustment based on input confidence scores\cite{li2023seenn,li2023unleashing}, stage-wise network partitioning with progressively shorter temporal horizons\cite{ding2024shrinking}, and stochastic latency training\cite{anumasa2024enhancing} have all been explored to enable early decision making.
In terms of memory optimization, online training approaches\cite{xiao2022online,meng2022training,jiang2024ndot} truncate or approximate the temporal gradient flow, preserving only essential state variables and thereby alleviating memory pressure. Other methods further reduce surrogate gradient computations from multi-timestep to single-timestep formulations\cite{suetake2023s3nn}, minimizing redundant temporal operations. Alternative frameworks restructure the training process to emphasize spatial rather than temporal backpropagation\cite{tan2021improved,meng2023towards}, effectively lowering the dependence on complete temporal sequences and improving storage efficiency during learning.
Architectural techniques have also been proposed to reduce memory overhead during training. Layer-wise state sharing\cite{kim2023sharing} enables memory reuse by allowing multiple layers to operate on common neuronal representations, thereby reducing the number of distinct activations that must be stored. 

In addition, reversible architectures in spiking neural networks\cite{zhang2024memory,zhao2024dr2net} employ invertible mappings to reconstruct forward activations during the backward pass, effectively eliminating the need to store intermediate temporal states.
For training acceleration, the SSF method \cite{wang2023ssf} stabilizes gradient propagation by averaging activations over time, mitigating temporal fluctuations in the gradient flow.T-RevSNN improves training efficiency by disabling the temporal dynamics of most spiking neurons and retaining only critical pathways for sequential updates, while employing multi-level reversible interactions to preserve model performance\cite{hu2024high}. In addition, A parallel training approach for SNNs is introduced, achieving constant time complexity across timesteps and enabling efficient large-scale learning.
However, most existing methods rely on sequential layer-wise computation, limiting parallelism and latency costs. ParaRevSNN addresses this issue by enabling parallel training across layers through reversible blocks with explicit reconstruction rules, reducing memory usage and improving training speed without sacrificing accuracy.
\section{Methodology}
\begin{figure*}[t]
\centering
\includegraphics[width=1.0\textwidth]{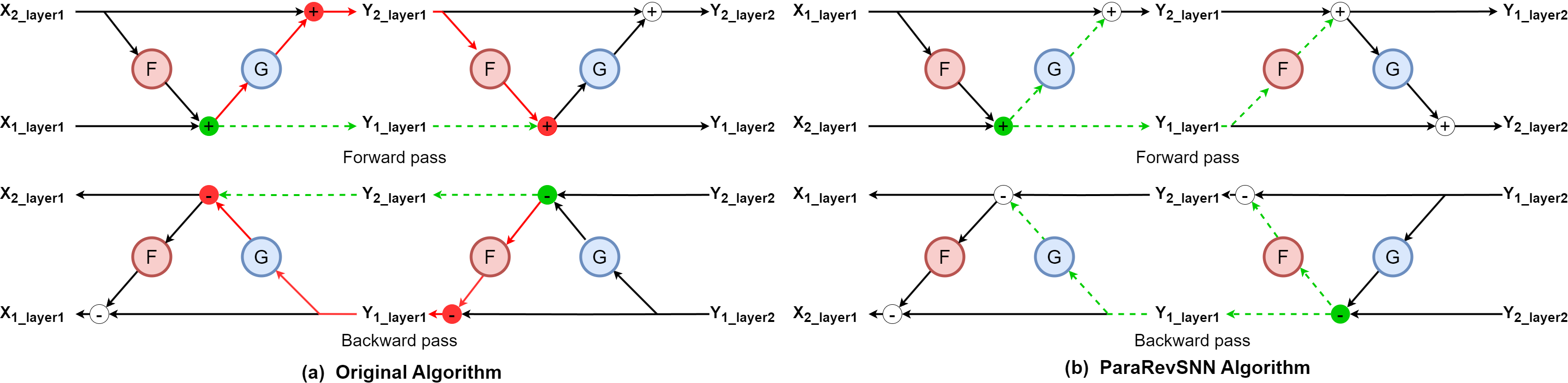} 
\caption{Comparison between the original reversible SNN\cite{zhang2024memory} and the proposed ParaRevSNN. (a) The red paths indicate sequential dependencies, where the computation of $F$ in the next layer depends on the output of $G$ in the current layer, restricting inter-layer parallelism. (b) The green paths in ParaRevSNN illustrate decoupled computation, enabling parallel execution of $F$ and $G$ across blocks, thus improving computational efficiency.}
\label{fig:graph-compare}
\end{figure*}

\subsection{Limitations of Existing Spiking Reversible Blocks}

Recent works have introduced reversible architectures into SNNs to reduce memory consumption during training~\cite{gomez2017reversible,zhang2024memory}. These models adopt a two-stream reversible structure inspired by RevNet, and extend it to SNNs by treating reversibility along the spatial dimension while maintaining consistency in the temporal domain.

Specifically, at each time step \(t\), a standard spiking reversible block is formulated as:
\begin{equation}
\begin{aligned}
y_1 &= x_1 + F(x_2), \\
y_2 &= x_2 + G(y_1),
\end{aligned}
\label{eq:rev-block}
\end{equation}
where \(F\) and \(G\) denote spiking modules that can include convolutions, attention mechanisms, or MLPs, and \(x_1, x_2\) are the dual input streams. The backward pass avoids caching intermediate variables by recomputing them through a reverse process:
\begin{equation}
\begin{aligned}
x_2 &= y_2 - G(y_1), \\
x_1 &= y_1 - F(x_2).
\end{aligned}
\end{equation}

Although the original reversible spiking network design supports full recomputation during backpropagation and significantly reduces memory usage, the computation graph still exhibits critical sequential dependencies. 

Within each reversible block, the non-linear transformation $G(y_1)$ must wait for the output $y_1 = x_1 + F(x_2)$ to be computed. More importantly, in a multi-layer stacking scenario, the computation of $F(y_1)$ in the next layer cannot proceed until $G(y_1)$ in the current layer has been completed. 

This enforces a strict inter-layer dependency between $G$ and $F$, effectively serializing the execution of all $F$ and $G$ functions across the network. Such dependencies limit the degree of parallelism that can be exploited by hardware accelerators and inevitably increase the overall training and inference latency.

These limitations motivate us to redesign the reversible block structure to break such sequential constraints and enable efficient layer-wise parallel computation.

\subsection{ParaRevSNN: Enabling Layer-wise Parallelism}

To address the above limitation, we propose ParaRevSNN, a novel reversible spiking architecture that reorders the data flow between the two residual streams to enable inter-layer parallelism. Our design is formulated as:
\begin{equation}
\begin{aligned}
y_1 &= x_2 + F(x_1), \\
y_2 &= x_1 + G(y_1),
\end{aligned}
\label{eq:para-rev-block}
\end{equation}

By switching the inputs of \(F\) and \(G\), we preserve reversibility while unlocking parallelism opportunities. When stacking blocks sequentially, the next block takes \(y_1, y_2\) as inputs:
\begin{equation}
\begin{aligned}
y_{11} &= y_2 + F(y_1), \\
y_{22} &= y_1 + G(y_{11}).
\end{aligned}
\label{eq:para-rev-next}
\end{equation}

Crucially, once \(y_1\) is computed via Eq.~\eqref{eq:para-rev-block}, both \(G(y_1)\) in the current block and \(F(y_1)\) in the next block can be computed in parallel. This breaks the strict sequential dependency observed in Eq.~\eqref{eq:rev-block}, as illustrated in Figure~\ref{fig:graph-compare}.

\subsection{Residuality and Reversibility}

The reversible formulation of ParaRevSNN follows the structure of residual reversible networks, where the forward and inverse computations are defined as:
\begin{equation}
\begin{aligned}
x_1 &= y_2 - G(y_1), \\
x_2 &= y_1 - F(x_1).
\end{aligned}
\end{equation}
This design is based on residual blocks, where $F$ and $G$ serve as transformation functions with skip connections. Such residual structures not only facilitate gradient flow during training but also enable exact reversibility.

Similar to prior reversible SNNs, membrane potentials and spike states at each time step can be recomputed during the backward pass. As a result, ParaRevSNN maintains memory efficiency with a peak cost of $\mathcal{O}(T)$ instead of $\mathcal{O}(D \cdot T)$, where $D$ is the number of layers and $T$ the number of time steps.

\subsection{Parallel Reversible Spiking Residual Network}
\begin{figure*}[t]
\centering
\includegraphics[width=0.8\textwidth]{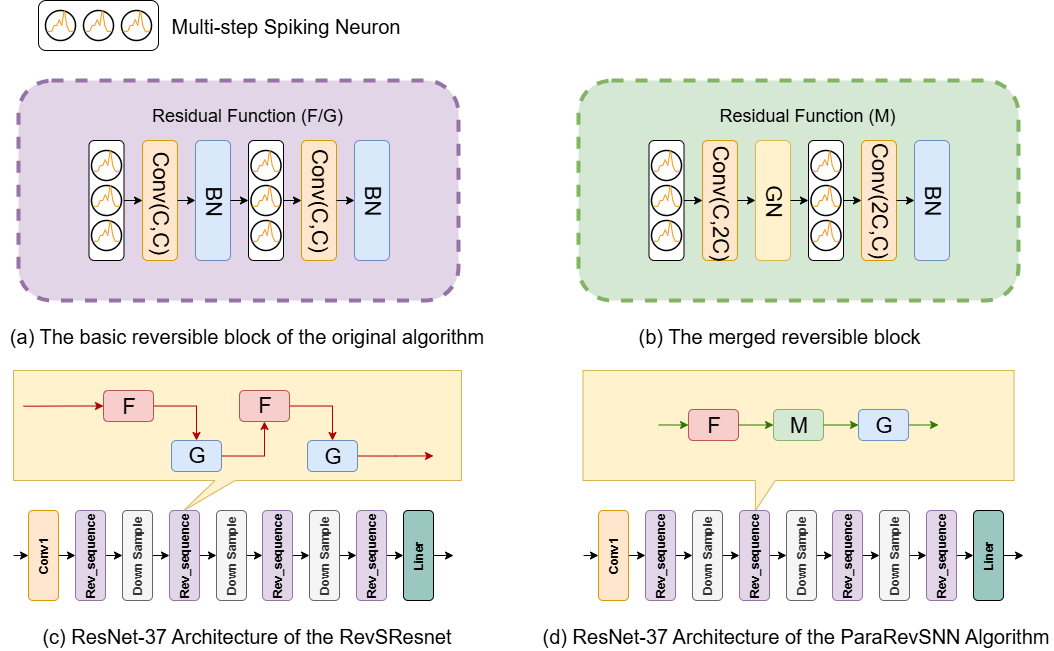} 
\caption{
Comparison between the original algorithm and the proposed ParaRevSNN Algorithm. 
(a) The structure of the original reversible block, which consists of two alternating activation-convolution-BN sequences. Our non-merged reversible blocks retain this architecture. 
(b) The proposed merged reversible block, designed as part of the efficient implementation of our ParaRevSNN. Compared with (a), it only differs in the number of channels and is used to compute $y_2$ in Eq.~\ref{eq:para-rev-block} and $y_{11}$ in Eq.~\ref{eq:para-rev-next}. 
(c) The RevSResNet-37 architecture constructed by the original algorithm, where the reversible layers are composed of two pairs of reversible blocks as shown in (a), connected in a serial manner. The detailed connection is illustrated in Fig.~\ref{fig:graph-compare} (a). 
(d) The ResNet-37 architecture constructed by our proposed ParaRevSNN algorithm, where the reversible layers are composed of both types of reversible blocks shown in (a) and (b).
}

\label{fig:basic_blcok}
\end{figure*}

In our proposed ParaRevSNN, we follow the basic reversible block structure proposed in~\cite{zhang2024memory}, as shown in Figure~\ref{fig:graph-compare}(a). As previously analyzed, in our parallel reversible architecture, the functions \( G(y_1) \) (Eq.~\ref{eq:para-rev-block}) and \( F(y_1) \) (Eq.~\ref{eq:para-rev-next}) from adjacent reversible blocks share the same input and exhibit no computational dependency. 

To further improve parallel execution efficiency, we structurally fuse the two residual functions into a unified residual module \( M(y_1) \). The overall structure follows a two-stage design comprising activation, convolution, and normalization layers. Specifically, the layout is: activation, followed by a convolution layer, group normalization (GN), another activation, a second convolution, and batch normalization (BN). While batch normalization is commonly used in spiking neural networks, we adopt GN in the middle normalization stage to better capture spatial and channel-wise statistical characteristics. This choice empirically improves training stability and performance in deep spiking architectures, especially under small batch sizes or with highly dynamic spike patterns.
To preserve functional equivalence with the original design, we adjust the number of channels in both convolutional layers. In particular, the second convolution is configured to emulate the additive structure of \( y_{11} \) in Eq.~\ref{eq:para-rev-next}. The resulting fused reversible block is illustrated in Figure~\ref{fig:graph-compare}(b). This design maintains the reversibility constraints while enhancing inter-block parallelism, making it more suitable for high-performance deployment.

\section{Experiments and results}
\begin{table}[htbp]
\centering
\renewcommand{\arraystretch}{1.5}
\begin{tabular}{c|c}
\hline\hline
Total layers & $N = 5 + 4 \times \sum n_i$ \\
\hline
conv1 & $3 \times 3 , 128$ \\ 
\hline
reversible sequence 1 & 
$\left(
  \begin{array}{l}
  3 \times 3,\ 64 \\
  3 \times 3,\ 64
  \end{array}
\right) \times 2 \times n_1$
 \\
\hline
reversible sequence 2 & $\left(
  \begin{array}{l}
  3 \times 3,\ 128 \\
  3 \times 3,\ 128
  \end{array}
\right)^{*} \times 2 \times n_2$ \\
\hline
reversible sequence 3 & $\left(
  \begin{array}{l}
  3 \times 3,\ 256 \\
  3 \times 3,\ 256
  \end{array}
\right)^{*} \times 2 \times n_3$ \\
\hline
reversible sequence 4 & $\left(
  \begin{array}{l}
  3 \times 3,\ 448 \\
  3 \times 3,\ 448
  \end{array}
\right)^{*} \times 2 \times n_4$ \\
\hline
  & average pool,fc,softmax \\
\hline\hline
\end{tabular}
\caption{Architectures of RevSResNet. The stride of \texttt{conv1} is set to 2 for downsampling. * indicates that a downsampling block is placed at the beginning of the reversible sequence. \(N\) represents the total number of layers. When the number of layers in a reversible group satisfies \(n_i \geq 2\), the proposed ParaRevSNN enables concurrent execution of these layers to improve computational efficiency.
}
\label{tab:layer_num}
\end{table}

\begin{table*}[htbp]
\centering
\tiny
\renewcommand{\arraystretch}{1.0}
\begin{tabular}{c|c|c|c|c|c|c|c|c|c|c|c}
\hline \hline
\multirow{3}{*}{Methods} & \multirow{3}{*}{Arch} & \multicolumn{2}{c|}{Efficiency} & \multicolumn{4}{c|}{CIFAR10} & \multicolumn{4}{c}{CIFAR100}\\
\cline{3-12}
 &  & Para & Time  & Top1     & Train     & Inference     & Mem       & Top1     & Train     & Inference & Mem          \\
 &  & (M)  & Step  & acc(\%)  & time(h)   & time(us/img)      & (MB/img)  & acc(\%)  & time(h)   & time(us/img)  & (MB/img)   \\
\hline
 STBP-tdBN &\multirow{2}{*}{ResNet19}  & \multirow{2}{*}{12.63}  & \multirow{2}{*}{4} & \multirow{2}{*}{92.92} & \multirow{2}{*}{-} & \multirow{2}{*}{-}  & \multirow{2}{*}{-} & \multirow{2}{*}{70.86} & \multirow{2}{*}{-} & \multirow{2}{*}{-}   &   \multirow{2}{*}{-} \\
 \cite{zheng2021going}&                            &                         &                    &   &   &       &   &   &  & \\
\hline
 TET &\multirow{2}{*}{ResNet19}  & \multirow{2}{*}{12.63}  & \multirow{2}{*}{4} & \multirow{2}{*}{94.44} & \multirow{2}{*}{-} & \multirow{2}{*}{-}  & \multirow{2}{*}{-} & \multirow{2}{*}{74.47} & \multirow{2}{*}{-} & \multirow{2}{*}{-}  &   \multirow{2}{*}{-}\\
 \cite{deng2022temporal}&                            &                         &                    &   &   &       &   &   & & &\\
 \hline
 OTTT &\multirow{2}{*}{VGG}  & \multirow{2}{*}{9.2}  & \multirow{2}{*}{6} & \multirow{2}{*}{93.58} & \multirow{2}{*}{-} & \multirow{2}{*}{-}  & \multirow{2}{*}{-} & \multirow{2}{*}{71.11} & \multirow{2}{*}{-} & \multirow{2}{*}{-}  &   \multirow{2}{*}{-}  \\
 \cite{xiao2022online}&                            &                         &                    &   &   &       &   &   & & &\\
  \hline
 SLTT &\multirow{2}{*}{ResNet18}  & \multirow{2}{*}{-}  & \multirow{2}{*}{20} & \textbf{\multirow{2}{*}{95.40}} & \multirow{2}{*}{-} & \multirow{2}{*}{-}  & \multirow{2}{*}{-} & \multirow{2}{*}{\textbf{78.50}} & \multirow{2}{*}{-} & \multirow{2}{*}{-}  &   \multirow{2}{*}{-} \\
 \cite{meng2023towards}&                            &                         &                    &   &   &       &   &   &  &  &\\
 \hline
  EfficientLIF-Net &\multirow{2}{*}{ResNet-19}  & \multirow{2}{*}{-}  & \multirow{2}{*}{5} & \multirow{2}{*}{91.92} & \multirow{2}{*}{-} & \multirow{2}{*}{-}  & \multirow{2}{*}{-} & \multirow{2}{*}{70.01} & \multirow{2}{*}{-} & \multirow{2}{*}{-}  &   \multirow{2}{*}{-}  \\
 \cite{kim2023sharing}&                            &                         &                    &   &   &       &   &   & & &\\
  \hline \hline
  MS ResNet &\multirow{2}{*}{ResNet18}  & \multirow{2}{*}{11.22}  & \multirow{2}{*}{4}& \multirow{2}{*}{94.33} & \multirow{2}{*}{3.56}  & \multirow{2}{*}{\textbf{12.19}} & \multirow{2}{*}{58.88} & \multirow{2}{*}{75.14} & \multirow{2}{*}{4.4} & \multirow{2}{*}{\textbf{11.88}}  & \multirow{2}{*}{60.16} \\
 \cite{hu2024high}&                            &                         &                    &   &   &       &   &   &  &   &\\
 \hline
   RevSResNet &\multirow{2}{*}{ResNet21}  & \multirow{2}{*}{11.05}  & \multirow{2}{*}{4} & \textbf{\multirow{2}{*}{94.57}} & \multirow{2}{*}{$3.63$} & \multirow{2}{*}{13.44} & \multirow{2}{*}{$32.41\downarrow\times1.82$} & \multirow{2}{*}{\textbf{75.71}}& \multirow{2}{*}{$3.55$} & \multirow{2}{*}{14.38} &  \multirow{2}{*}{$\mathbf{32.41}\downarrow\times1.86$} \\
 \cite{zhang2024memory}&                            &                         &                    &   &   &       &   &   &  &   &\\
 \hline
    ParaRevSNN &\multirow{2}{*}{ResNet21}  & \multirow{2}{*}{11.05}  & \multirow{2}{*}{4} & \multirow{2}{*}{94.47} & \multirow{2}{*}{$\mathbf{3.55}$} & \multirow{2}{*}{12.81}
 & \multirow{2}{*}{$\mathbf{32.41}\downarrow\times1.82$} & \multirow{2}{*}{75.55}& \multirow{2}{*}{$\mathbf{3.52}$} & \multirow{2}{*}{14.06} & \multirow{2}{*}{$\mathbf{32.41}\downarrow\times1.86$} \\
 (ours)&                            &                         &                    &   &   &       &   &   &  &   &\\
 \hline \hline
  MS ResNet &\multirow{2}{*}{ResNet34}  & \multirow{2}{*}{21.33}  & \multirow{2}{*}{4}& \multirow{2}{*}{94.82} & \multirow{2}{*}{\textbf{4.40}} & \multirow{2}{*}{17.81} & \multirow{2}{*}{103.59} & \multirow{2}{*}{75.39} & \multirow{2}{*}{\textbf{5.28}} & \multirow{2}{*}{18.13} & \multirow{2}{*}{98.63} \\
 \cite{hu2024high}&                            &                         &                    &   &   &       &   &   &   &   &\\
 \hline
   RevSResNet &\multirow{2}{*}{ResNet37}  & \multirow{2}{*}{21.16}  & \multirow{2}{*}{4} & \textbf{\multirow{2}{*}{95.04}} & \multirow{2}{*}{$6.45$} & \multirow{2}{*}{20.00} & \multirow{2}{*}{$\mathbf{38.13}\downarrow\times140$} & \multirow{2}{*}{\textbf{76.22}}& \multirow{2}{*}{$6.56$}  & \multirow{2}{*}{20.31} & \multirow{2}{*}{$\mathbf{38.13}\downarrow\times2.59$} \\
 \cite{zhang2024memory}&                            &                         &                    &   &   &       &   &   &  &   &\\
 \hline
    ParaRevSNN &\multirow{2}{*}{ResNet37}  & \multirow{2}{*}{21.16}  & \multirow{2}{*}{4} & \multirow{2}{*}{94.92} & \multirow{2}{*}{$5.43$} & \multirow{2}{*}{\textbf{17.50}} & \multirow{2}{*}{$44.00\downarrow\times2.35$} & \multirow{2}{*}{75.55}& \multirow{2}{*}{$5.40$} & \multirow{2}{*}{\textbf{17.19}} & \multirow{2}{*}{$44.00\downarrow\times2.24$} \\
 (ours)&                            &                         &                    &   &   &       &   &   &  &   &\\
\hline\hline
\end{tabular}
\caption{Comparison of Efficiency and Performance Across SNN Models on CIFAR10 and CIFAR100.}
\label{tab:static_datasets}
\end{table*}
In this section, we evaluate the proposed ParaRevSNN on two static image classification datasets, CIFAR10 and CIFAR100\cite{krizhevsky2009learning}, as well as two neuromorphic datasets, CIFAR10-DVS~\cite{li2017cifar10} and DVS128 Gesture\cite{amir2017low}. We compare different SNN architectures using multiple metrics, including the number of parameters, time steps, FLOPS, peak memory usage per image, and top-1 accuracy.The \textit{memory per image} is measured as the peak GPU memory occupied by a single image during training. To enable a fair comparison with non-reversible baselines, we align the model complexity (in terms of FLOPS) and the number of parameters as closely as possible.In addition, to assess the acceleration benefits of our approach, we also report the total training time.

\begin{table*}[htbp]
\centering
\tiny
\renewcommand{\arraystretch}{1.0}
\begin{tabular}{c|c|c|c|c|c|c|c|c|c|c|c|c}
\hline \hline
\multirow{3}{*}{Methods} & \multirow{3}{*}{Arch} &  \multirow{2}{*}{FLOPS} & \multicolumn{5}{c|}{CIFAR10-DVS} & \multicolumn{5}{c}{DVS128 Gesture}\\
\cline {4-13}
 &  &      & Time  & Top1 & Train    & Inference & Mem      & Time   & Top1 & Train   &Inference   & Mem        \\
 &  & (G)  & Step  & acc  & time(h)  & time(us/img)  & (MB/img) & Step   & acc  & time(h) &time(us/img)    & (MB/img)   \\
\hline
 T-RevSNN &\multirow{2}{*}{ResNet18}  & \multirow{2}{*}{-}  & \multirow{2}{*}{16} & \multirow{2}{*}{79.20}  & \multirow{2}{*}{-} & \multirow{2}{*}{-} & \multirow{2}{*}{18.9}  & \multirow{2}{*}{16} & \multirow{2}{*}{97.9} & \multirow{2}{*}{-} & \multirow{2}{*}{-} & \multirow{2}{*}{18.9} \\
 \cite{hu2024high}&                            &                         &                    &   &   &       &   &   & & &  \\
\hline
 SEW ResNet &\multirow{2}{*}{ResNet19}  & \multirow{2}{*}{-}  & \multirow{2}{*}{16} & \multirow{2}{*}{74.40} & \multirow{2}{*}{-}  & \multirow{2}{*}{-} & \multirow{2}{*}{81.50} & \multirow{2}{*}{16} & \multirow{2}{*}{97.20} & \multirow{2}{*}{-} & \multirow{2}{*}{-} & \multirow{2}{*}{81.50} \\
 \cite{fang2021deep}&                            &                         &                    &   &   &       &   &   &&   &\\
 \hline
 OTTT &\multirow{2}{*}{VGG}  & \multirow{2}{*}{-}  & \multirow{2}{*}{10} & \multirow{2}{*}{76.6} & \multirow{2}{*}{-} & \multirow{2}{*}{-}  & \multirow{2}{*}{-} & \multirow{2}{*}{20} & \multirow{2}{*}{96.90} & \multirow{2}{*}{-} & \multirow{2}{*}{-} & \multirow{2}{*}{-} \\
 \cite{xiao2022online}&                            &                         &                    &   &   &       &   &   &  &   &\\
  \hline
 SLTT &\multirow{2}{*}{ResNet18}  & \multirow{2}{*}{-}  & \multirow{2}{*}{10} & \multirow{2}{*}{77.20} & \multirow{2}{*}{-} & \multirow{2}{*}{-}  & \multirow{2}{*}{-} & \multirow{2}{*}{10} & \multirow{2}{*}{97.9} & \multirow{2}{*}{-}  & \multirow{2}{*}{-} & \multirow{2}{*}{-}  \\
 \cite{meng2023towards}&                            &                         &                    &   &   &       &   &   & &   &\\
 \hline
  FPT &\multirow{2}{*}{VGG11}  & \multirow{2}{*}{-}  & \multirow{2}{*}{10} & \multirow{2}{*}{\textbf{85.5}} & \multirow{2}{*}{-} & \multirow{2}{*}{-}  & \multirow{2}{*}{-} & \multirow{2}{*}{20} & \multirow{2}{*}{\textbf{98.38}} & \multirow{2}{*}{-}  & \multirow{2}{*}{-} & \multirow{2}{*}{-}  \\
 \cite{feng2025efficient}&                            &                         &                    &   &   &       &   &   & &   &\\
  \hline \hline
  MS ResNet &\multirow{2}{*}{ResNet20}  & \multirow{2}{*}{0.43}  & \multirow{2}{*}{10}& \multirow{2}{*}{73.80} & \multirow{2}{*}{\textbf{0.57}} & \multirow{2}{*}{137} & \multirow{2}{*}{28.50} & \multirow{2}{*}{10} & \multirow{2}{*}{\textbf{94.44}} & \multirow{2}{*}{\textbf{0.18}} & \multirow{2}{*}{\textbf{141}} & \multirow{2}{*}{28.50} \\
 \cite{hu2024high}&                            &                         &                    &   &   &       &   &   &  &   & \\
 \hline
   RevSResNet &\multirow{2}{*}{ResNet24}  & \multirow{2}{*}{0.43}  & \multirow{2}{*}{10} & \multirow{2}{*}{\textbf{75.50}} & \multirow{2}{*}{$0.93$}  & \multirow{2}{*}{151} & \multirow{2}{*}{$\mathbf{21.76}\downarrow\times1.31$} & \multirow{2}{*}{10} & \multirow{2}{*}{93.06} & \multirow{2}{*}{$0.28$}  & \multirow{2}{*}{156} & \multirow{2}{*}{$\mathbf{21.76}\downarrow\times1.31$} \\
 \cite{zhang2024memory}&                            &                         &                    &   &   &       &   &   &  &   & \\
 \hline
    ParaRevSNN &\multirow{2}{*}{ResNet24}  & \multirow{2}{*}{0.43}  & \multirow{2}{*}{10} & \multirow{2}{*}{\textbf{75.50}} & \multirow{2}{*}{$0.83$}  & \multirow{2}{*}{\textbf{150}} 
 & \multirow{2}{*}{$\mathbf{32.41}\downarrow\times1.14$} & \multirow{2}{*}{10} & \multirow{2}{*}{\textbf{94.44}}& \multirow{2}{*}{${0.27}$} & \multirow{2}{*}{149} &\multirow{2}{*}{$23.50\downarrow\times1.21$} \\
 (ours)&                            &                         &                    &   &   &       &   &   &  &   &\\
 \hline \hline
  MS ResNet &\multirow{2}{*}{ResNet20}  & \multirow{2}{*}{0.67}  & \multirow{2}{*}{16}& \multirow{2}{*}{77.70} & \multirow{2}{*}{0.63}  & \multirow{2}{*}{\textbf{172}} & \multirow{2}{*}{43.06} & \multirow{2}{*}{16} & \multirow{2}{*}{\textbf{96.88}} & \multirow{2}{*}{\textbf{0.23}} & \multirow{2}{*}{\textbf{173}}  & \multirow{2}{*}{43.06}\\
 \cite{hu2024high}&                            &                         &                    &   &   &       &   &   &&   &\\
 \hline
   RevSResNet &\multirow{2}{*}{ResNet24}  & \multirow{2}{*}{0.43}  & \multirow{2}{*}{16} & \multirow{2}{*}{75.70} & \multirow{2}{*}{$1.00$}  & \multirow{2}{*}{182}  & \multirow{2}{*}{$\mathbf{29.13}\downarrow\times1.48$} & \multirow{2}{*}{16} & \multirow{2}{*}{95.83} & \multirow{2}{*}{$0.31$}  & \multirow{2}{*}{$178$}  & \multirow{2}{*}{$\mathbf{29.13}\downarrow\times2.72$} \\
 \cite{zhang2024memory}&                            &                         &                    &   &   &       &   &   & &   &\\
 \hline
    ParaRevSNN &\multirow{2}{*}{ResNet24}  & \multirow{2}{*}{0.69}  & \multirow{2}{*}{16} &  \textbf{\multirow{2}{*}{75.80}} & \multirow{2}{*}{$\mathbf{0.88}$}  & \multirow{2}{*}{178} & \multirow{2}{*}{$36.25\downarrow\times1.19$} & \multirow{2}{*}{16} & \multirow{2}{*}{96.53}& \multirow{2}{*}{0.27} & \multirow{2}{*}{175}  & \multirow{2}{*}{$36.25\downarrow\times2.35$} \\
 (ours)&                            &                         &                    &   &   &       &   &   &  &   & \\
\hline\hline
\end{tabular}
\caption{Comparison of Efficiency and Performance Across SNN Models on CIFAR10-DVS and DVS128 Gesture.}
\label{tab:neo_datasets}
\end{table*}

\begin{table*}[htbp]
\scriptsize
\centering
\renewcommand{\arraystretch}{1.0}
\begin{tabular}{@{} c | c | c | c | c | c | c | c | c @{}}
\hline\hline
Dataset & Network & Blocks & Channels & Neuron & Epocch & Learning Rate & Optimizer & Batch Size \\ 
\hline
\multirow{2}{*}{Static} 
  & MS ResNet18     & 2-2-2-2     & 64-128-256-512   & IF & 200 & 0.1 & SGD & 32  \\
  & RevSResNet21    & 1-1-1-1     & 64-128-256-448   & IF & 200 & 0.1 & SGD & 32  \\
\hline
\multirow{2}{*}{Static} 
  & MS ResNet34     & 3-4-6-3     & 64-128-256-512   & IF & 200 & 0.1 & SGD & 32  \\
  & RevSResNet37    & 2-2-2-2     & 64-128-256-448   & IF & 200 & 0.1 & SGD & 32  \\
\hline
\multirow{2}{*}{Neuromorphic} 
  & MS ResNet20     & 3-3-3       & 16-32-64         & LIF & 100(CIFAR10-DVS)/200(DVS128 Gesture) & 0.001 & AdamW & 16  \\
  & RevSResNet24    & 1-2-2       & 16-32-48         & LIF & 100(CIFAR10-DVS)/200(DVS128 Gesture) & 0.001 & AdamW & 16  \\
\hline\hline
\end{tabular}
\caption{Model size across static and neuromorphic datasets.}
\label{tab:experiment-settings}
\end{table*}

\subsection{Experiment Settings}
Our experiments are implemented based on the official open-source code provided by prior works\cite{zhang2024memory}. To ensure fair comparison, we follow the original training protocols, including data preprocessing, learning rate schedules, and optimization strategies. All experiments are conducted on a single NVIDIA RTX 4090 GPU and an Intel(R) Xeon(R) Gold 5418Y CPU using PyTorch with automatic mixed precision enabled. The detailed experimental configurations, including the network architecture, spiking neuron models, number of epochs, learning rate, optimizer, and batch size, are summarized in Table~\ref{tab:experiment-settings}.

\subsection{Evaluation on Static Datasets}

We evaluate the proposed ParaRevSNN on two widely used static image classification benchmarks: CIFAR10 and CIFAR100. Both datasets consist of 60,000 RGB images of size $32 \times 32$, with 50,000 images for training and 10,000 for testing. CIFAR10 contains 10 coarse-grained classes, while CIFAR100 provides a more challenging setting with 100 fine-grained classes.

To assess the performance of the proposed parallel reversible architecture, we conduct experiments under two different network depths: a 21-layer shallow network and a 37-layer deeper network. The 21-layer setting adopts a sequential structure (ResNet21), whereas the 37-layer setting introduces our multi-stage parallel design (ResNet37).

In the 21-layer configuration, ParaRevSNN achieves comparable accuracy to the non-parallel baseline RevSResNet, with top-1 accuracies of 94.47\% and 94.57\% on CIFAR10, and 75.55\% versus 75.71\% on CIFAR100, showing that parallel computation does not compromise performance in shallower networks. ParaRevSNN also reduces total training time to 3.55 hours (CIFAR10) and 3.52 hours (CIFAR100), with a slight improvement in inference latency (12.81 $\mu$s vs. 13.44 $\mu$s per image). In the deeper 37-layer setting, these benefits become more pronounced: training time decreases by 15.8\% (from 6.45 to 5.43 hours on CIFAR10), inference time per image drops from 20.00 $\mu$s to 17.50 $\mu$s, and memory usage is significantly reduced by 2.35$\times$ (from 44.00 MB to 18.75 MB per image). This demonstrates that ParaRevSNN effectively exploits parallelism in deeper networks, yielding substantial efficiency gains without sacrificing accuracy.

\subsubsection{Accuracy-Efficiency Trade-off.} 
While ParaRevSNN demonstrates clear advantages in training time, inference speed, and memory usage, a slight drop in accuracy is observed in deeper models. On CIFAR10, ParaRevSNN achieves 94.92\% accuracy, compared to 95.04\% from RevSResNet. On CIFAR100, the accuracy drops from 76.22\% to 75.55\%. We hypothesize that this drop is due to reduced spatial information interaction within parallel branches, which may weaken long-range dependencies that sequential architectures naturally capture. Nonetheless, the accuracy loss is marginal and acceptable, especially considering the substantial efficiency gains. In latency- or memory-critical applications, ParaRevSNN offers a compelling trade-off.

\subsection{Experiments on Neuromorphic Datasets}

We further evaluate our proposed ParaRevSNN on two widely-used neuromorphic vision benchmarks: CIFAR10-DVS and DVS128 Gesture. CIFAR10-DVS is a dynamic vision sensor version of CIFAR10, containing 10 object classes with event-based data captured in real-world conditions. DVS128 Gesture consists of 11 dynamic hand gestures recorded using a DVS camera, providing a more complex temporal pattern recognition challenge.

\subsubsection{Training Performance and Efficiency Analysis}
Table~\ref{tab:neo_datasets} shows ParaRevSNN achieves comparable or better accuracy on both datasets. On CIFAR10-DVS, it reaches 75.50\% accuracy with 0.83 hours training, surpassing RevSResNet in efficiency and memory use. On DVS128 Gesture, ParaRevSNN matches MS-ResNet’s 94.44\% accuracy while reducing memory by $1.21\times$. In the deeper 16-step setting, it maintains strong accuracy (75.80\% and 96.53\%) with further reductions in training time and memory, lowering memory usage by $2.35\times$ on DVS128 and $1.19\times$ on CIFAR10-DVS.

\subsubsection{Inference Speed and Inference Efficiency.} 
Our model also demonstrates improved inference speed across both datasets. On CIFAR10-DVS, ParaRevSNN reaches an inference latency of 150~$\mu$s per image in the 10-step setting, better than RevSResNet (151~$\mu$s) and significantly faster than MS-ResNet (172~$\mu$s in the 16-step setting). These results highlight that ParaRevSNN enables faster inference while maintaining performance.


\subsection{Ablation Studies}
\subsubsection{Impact of Spiral Reversibility on Accuracy}
In our proposed reversible architecture, the spatial information flow forms a spiral-like dependency when only one $(F, G)$ pair is present. Although the reversed reconstruction of features remains feasible after spatial downsampling, an odd number of reversible pairs may disrupt structural symmetry, potentially degrading the final accuracy.
To investigate this phenomenon, we design a 25-layer network containing 5 reversible $(F, G)$ pairs. Among them, two pairs are deliberately placed between different downsampling layers to create varied spatial information flow patterns. All models are evaluated on the CIFAR10 dataset to assess the effect of reversible pair placement and the parity (even vs. odd) of the total number of reversible blocks.
As shown in Table~\ref{tab:rev_struct_compare}, different arrangements lead to slight variations in classification accuracy. These results suggest that structural symmetry and the configuration of reversible modules can influence performance, even when the total depth remains constant.

\begin{table}[htbp]
\scriptsize
\centering
\begin{tabular}{c c c c }
\hline
\textbf{Network} & \textbf{Architecture} & \textbf{Original Accuracy (\%)} & \textbf{Our Accuracy (\%)} \\
\hline
Resnet25-1 & [2,1,1,1] & 94.51 & 94.40 \\
Resnet25-2 & [1,2,1,1] & 94.82 & 94.70 \\
Resnet25-3 & [1,1,2,1] & 94.58 & 94.58 \\
Resnet25-4 & [1,1,1,2] & 94.55 & 94.67 \\
\hline
\end{tabular}
\caption{Comparison of Accuracy Between Different Network Structures}
\label{tab:rev_struct_compare}
\end{table}

\begin{figure}[t]
\centering
\includegraphics[width=1.0\linewidth]{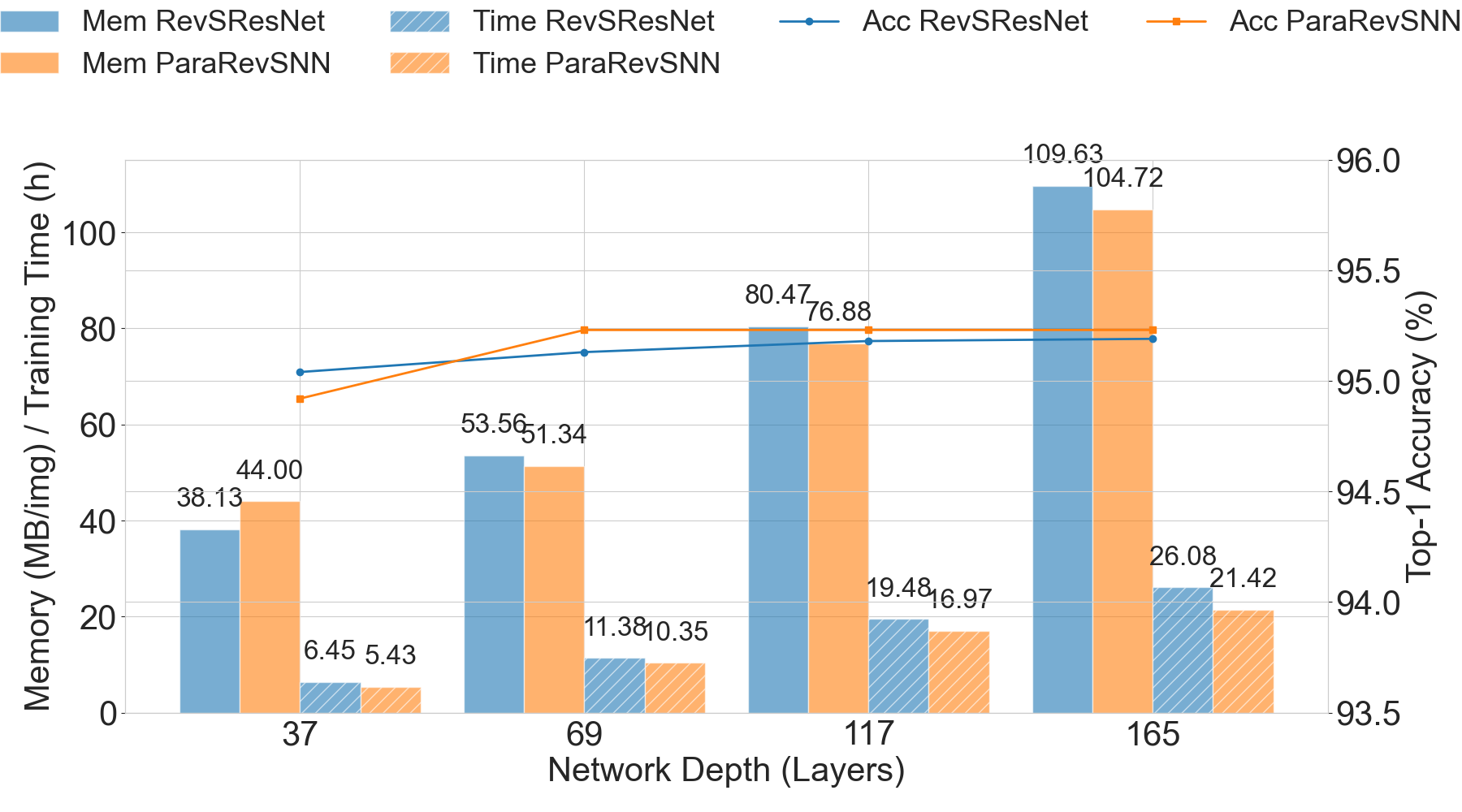}
\caption{Comparison of memory usage (MB/img), training time (h), and accuracy (\%) between ParaRevSNN and RevSResNet at different network depths, demonstrating how efficiency and performance evolve as network depth increases.}
\label{fig:Comparison of deeper}
\end{figure}

\subsubsection{Effectiveness of ParaRevSNN on Deeper Networks}

To assess the scalability and efficiency of ParaRevSNN, we evaluated it alongside the baseline RevSResNet on CIFAR10 using networks with depths of 37, 69, 117, and 165 layers. The comparison focuses on memory consumption, training time, and classification accuracy, as illustrated in Figure~\ref{fig:Comparison of deeper}.

The memory usage of ParaRevSNN is slightly higher than RevSResNet at shallow depths (e.g., 44 MB vs. 38.13 MB at 37 layers), likely due to parallelization overhead. However, as network depth increases, ParaRevSNN exhibits better memory efficiency, consuming less memory at 165 layers (104.72 MB vs. 109.63 MB), demonstrating favorable scaling characteristics.

Training time consistently benefits from the parallel design of ParaRevSNN across all depths, with acceleration more pronounced for deeper networks. For example, training time decreases from 6.45 hours to 5.43 hours at 37 layers, and from 26.08 hours to 21.42 hours at 165 layers, yielding approximately 18\% speedup at the deepest setting.

Importantly, the top-1 accuracy of ParaRevSNN remains on par or slightly better than the baseline at deeper depths (95.23\% vs. 95.19\% at 165 layers), confirming that efficiency gains do not come at the cost of performance.The results validate the effectiveness and scalability of ParaRevSNN, particularly for deeper reversible spiking neural networks where both memory savings and training acceleration are critical.

\section{Conclusion}
In this work, we proposed ParaRevSNN, a parallel reversible spiking neural network architecture designed to overcome the sequential dependency bottleneck in traditional reversible SNNs. By enabling inter-block parallelism, ParaRevSNN achieves significant improvements in training efficiency and memory utilization, especially as network depth increases. Experimental results on CIFAR10 demonstrate that ParaRevSNN consistently reduces training time and memory consumption while maintaining competitive or even slightly improved accuracy compared to the baseline RevSResNet. These advantages make ParaRevSNN a promising approach for scaling reversible spiking neural networks to deeper architectures, facilitating more efficient training and deployment in resource-constrained environments. Future work includes extending ParaRevSNN to larger datasets and exploring hardware acceleration to further leverage its parallelism potential.

\bibliography{aaai2026}

\begin{thebibliography}{40}
\providecommand{\natexlab}[1]{#1}

\bibitem[{Akopyan et~al.(2015)Akopyan, Sawada, Cassidy, Alvarez-Icaza, Arthur, Merolla, Imam, Nakamura, Datta, Nam et~al.}]{akopyan2015truenorth}
Akopyan, F.; Sawada, J.; Cassidy, A.; Alvarez-Icaza, R.; Arthur, J.; Merolla, P.; Imam, N.; Nakamura, Y.; Datta, P.; Nam, G.-J.; et~al. 2015.
\newblock Truenorth: Design and tool flow of a 65 mw 1 million neuron programmable neurosynaptic chip.
\newblock \emph{IEEE transactions on computer-aided design of integrated circuits and systems}, 34(10): 1537--1557.

\bibitem[{Amir et~al.(2017)Amir, Taba, Berg, Melano, McKinstry, Di~Nolfo, Nayak, Andreopoulos, Garreau, Mendoza et~al.}]{amir2017low}
Amir, A.; Taba, B.; Berg, D.; Melano, T.; McKinstry, J.; Di~Nolfo, C.; Nayak, T.; Andreopoulos, A.; Garreau, G.; Mendoza, M.; et~al. 2017.
\newblock A low power, fully event-based gesture recognition system.
\newblock In \emph{Proceedings of the IEEE conference on computer vision and pattern recognition}, 7243--7252.

\bibitem[{Anumasa et~al.(2024)Anumasa, Mukhoty, Bojkovic, De~Masi, Xiong, and Gu}]{anumasa2024enhancing}
Anumasa, S.; Mukhoty, B.; Bojkovic, V.; De~Masi, G.; Xiong, H.; and Gu, B. 2024.
\newblock Enhancing training of spiking neural network with stochastic latency.
\newblock In \emph{Proceedings of the AAAI Conference on Artificial Intelligence}, volume~38, 10900--10908.

\bibitem[{Aydin et~al.(2024)Aydin, Gehrig, Gehrig, and Scaramuzza}]{aydin2024hybrid}
Aydin, A.; Gehrig, M.; Gehrig, D.; and Scaramuzza, D. 2024.
\newblock A hybrid ANN-SNN architecture for low-power and low-latency visual perception.
\newblock In \emph{Proceedings of the IEEE/CVF Conference on Computer Vision and Pattern Recognition}, 5701--5711.

\bibitem[{Bojkovi{\'c}, Wu, and Gu(2025)}]{bojkovic2025temporal}
Bojkovi{\'c}, V.; Wu, X.; and Gu, B. 2025.
\newblock Temporal Misalignment in ANN-SNN Conversion and Its Mitigation via Probabilistic Spiking Neurons.
\newblock \emph{arXiv preprint arXiv:2502.14487}.

\bibitem[{Br{\"u}gger, Baumgartner, and Konukoglu(2019)}]{brugger2019partially}
Br{\"u}gger, R.; Baumgartner, C.~F.; and Konukoglu, E. 2019.
\newblock A partially reversible U-Net for memory-efficient volumetric image segmentation.
\newblock In \emph{International conference on medical image computing and computer-assisted intervention}, 429--437. Springer.

\bibitem[{Davies et~al.(2018)Davies, Srinivasa, Lin, Chinya, Cao, Choday, Dimou, Joshi, Imam, Jain et~al.}]{davies2018loihi}
Davies, M.; Srinivasa, N.; Lin, T.-H.; Chinya, G.; Cao, Y.; Choday, S.~H.; Dimou, G.; Joshi, P.; Imam, N.; Jain, S.; et~al. 2018.
\newblock Loihi: A neuromorphic manycore processor with on-chip learning.
\newblock \emph{Ieee Micro}, 38(1): 82--99.

\bibitem[{Deng et~al.(2020)Deng, Wu, Hu, Liang, Ding, Li, Zhao, Li, and Xie}]{deng2020rethinking}
Deng, L.; Wu, Y.; Hu, X.; Liang, L.; Ding, Y.; Li, G.; Zhao, G.; Li, P.; and Xie, Y. 2020.
\newblock Rethinking the performance comparison between SNNS and ANNS.
\newblock \emph{Neural networks}, 121: 294--307.

\bibitem[{Deng et~al.(2022)Deng, Li, Zhang, and Gu}]{deng2022temporal}
Deng, S.; Li, Y.; Zhang, S.; and Gu, S. 2022.
\newblock Temporal efficient training of spiking neural network via gradient re-weighting.
\newblock \emph{arXiv preprint arXiv:2202.11946}.

\bibitem[{Ding et~al.(2024)Ding, Zuo, Jing, He, and Xiao}]{ding2024shrinking}
Ding, Y.; Zuo, L.; Jing, M.; He, P.; and Xiao, Y. 2024.
\newblock Shrinking your timestep: Towards low-latency neuromorphic object recognition with spiking neural networks.
\newblock In \emph{Proceedings of the AAAI Conference on Artificial Intelligence}, volume~38, 11811--11819.

\bibitem[{Dinh, Krueger, and Bengio(2014)}]{dinh2014nice}
Dinh, L.; Krueger, D.; and Bengio, Y. 2014.
\newblock Nice: Non-linear independent components estimation.
\newblock \emph{arXiv preprint arXiv:1410.8516}.

\bibitem[{Fang et~al.(2021)Fang, Yu, Chen, Huang, Masquelier, and Tian}]{fang2021deep}
Fang, W.; Yu, Z.; Chen, Y.; Huang, T.; Masquelier, T.; and Tian, Y. 2021.
\newblock Deep residual learning in spiking neural networks.
\newblock \emph{Advances in Neural Information Processing Systems}, 34: 21056--21069.

\bibitem[{Feng et~al.(2025)Feng, Gao, Du, Shi, Zhao, Wu, and Miao}]{feng2025efficient}
Feng, W.; Gao, X.; Du, W.; Shi, H.; Zhao, P.; Wu, P.; and Miao, C. 2025.
\newblock Efficient Parallel Training Methods for Spiking Neural Networks with Constant Time Complexity.
\newblock \emph{arXiv preprint arXiv:2506.12087}.

\bibitem[{Gomez et~al.(2017)Gomez, Ren, Urtasun, and Grosse}]{gomez2017reversible}
Gomez, A.~N.; Ren, M.; Urtasun, R.; and Grosse, R.~B. 2017.
\newblock The reversible residual network: Backpropagation without storing activations.
\newblock \emph{Advances in neural information processing systems}, 30.

\bibitem[{Hu et~al.(2024)Hu, Yao, Qiu, Chou, Cai, Qiao, Tian, Xu, and Li}]{hu2024high}
Hu, J.; Yao, M.; Qiu, X.; Chou, Y.; Cai, Y.; Qiao, N.; Tian, Y.; Xu, B.; and Li, G. 2024.
\newblock High-Performance Temporal Reversible Spiking Neural Networks with $ O (L) $ Training Memory and $ O (1) $ Inference Cost.
\newblock \emph{arXiv preprint arXiv:2405.16466}.

\bibitem[{Jiang et~al.(2024)Jiang, De~Masi, Xiong, and Gu}]{jiang2024ndot}
Jiang, H.; De~Masi, G.; Xiong, H.; and Gu, B. 2024.
\newblock Ndot: Neuronal dynamics-based online training for spiking neural networks.
\newblock In \emph{Forty-first International Conference on Machine Learning}.

\bibitem[{Kim et~al.(2023)Kim, Li, Moitra, Yin, and Panda}]{kim2023sharing}
Kim, Y.; Li, Y.; Moitra, A.; Yin, R.; and Panda, P. 2023.
\newblock Sharing leaky-integrate-and-fire neurons for memory-efficient spiking neural networks.
\newblock \emph{Frontiers in Neuroscience}, 17: 1230002.

\bibitem[{Kim, Venkatesha, and Panda(2022)}]{kim2022privatesnn}
Kim, Y.; Venkatesha, Y.; and Panda, P. 2022.
\newblock Privatesnn: privacy-preserving spiking neural networks.
\newblock In \emph{Proceedings of the AAAI Conference on Artificial Intelligence}, volume~36, 1192--1200.

\bibitem[{Krizhevsky, Hinton et~al.(2009)}]{krizhevsky2009learning}
Krizhevsky, A.; Hinton, G.; et~al. 2009.
\newblock Learning multiple layers of features from tiny images.(2009).

\bibitem[{Lee, Delbruck, and Pfeiffer(2016)}]{lee2016training}
Lee, J.~H.; Delbruck, T.; and Pfeiffer, M. 2016.
\newblock Training deep spiking neural networks using backpropagation.
\newblock \emph{Frontiers in neuroscience}, 10: 508.

\bibitem[{Li, Jones, and Furber(2023)}]{li2023unleashing}
Li, C.; Jones, E.~G.; and Furber, S. 2023.
\newblock Unleashing the potential of spiking neural networks with dynamic confidence.
\newblock In \emph{Proceedings of the IEEE/CVF International Conference on Computer Vision}, 13350--13360.

\bibitem[{Li et~al.(2017)Li, Liu, Ji, Li, and Shi}]{li2017cifar10}
Li, H.; Liu, H.; Ji, X.; Li, G.; and Shi, L. 2017.
\newblock Cifar10-dvs: an event-stream dataset for object classification.
\newblock \emph{Frontiers in neuroscience}, 11: 244131.

\bibitem[{Li et~al.(2023)Li, Geller, Kim, and Panda}]{li2023seenn}
Li, Y.; Geller, T.; Kim, Y.; and Panda, P. 2023.
\newblock Seenn: Towards temporal spiking early exit neural networks.
\newblock \emph{Advances in Neural Information Processing Systems}, 36: 63327--63342.

\bibitem[{Maass(1997)}]{maass1997networks}
Maass, W. 1997.
\newblock Networks of spiking neurons: the third generation of neural network models.
\newblock \emph{Neural networks}, 10(9): 1659--1671.

\bibitem[{Mangalam et~al.(2022)Mangalam, Fan, Li, Wu, Xiong, Feichtenhofer, and Malik}]{mangalam2022reversible}
Mangalam, K.; Fan, H.; Li, Y.; Wu, C.-Y.; Xiong, B.; Feichtenhofer, C.; and Malik, J. 2022.
\newblock Reversible vision transformers.
\newblock In \emph{Proceedings of the IEEE/CVF Conference on Computer Vision and Pattern Recognition}, 10830--10840.

\bibitem[{Meng et~al.(2022)Meng, Xiao, Yan, Wang, Lin, and Luo}]{meng2022training}
Meng, Q.; Xiao, M.; Yan, S.; Wang, Y.; Lin, Z.; and Luo, Z.-Q. 2022.
\newblock Training high-performance low-latency spiking neural networks by differentiation on spike representation.
\newblock In \emph{Proceedings of the IEEE/CVF conference on computer vision and pattern recognition}, 12444--12453.

\bibitem[{Meng et~al.(2023)Meng, Xiao, Yan, Wang, Lin, and Luo}]{meng2023towards}
Meng, Q.; Xiao, M.; Yan, S.; Wang, Y.; Lin, Z.; and Luo, Z.-Q. 2023.
\newblock Towards memory-and time-efficient backpropagation for training spiking neural networks.
\newblock In \emph{Proceedings of the IEEE/CVF international conference on computer vision}, 6166--6176.

\bibitem[{Sander et~al.(2021)Sander, Ablin, Blondel, and Peyr{\'e}}]{sander2021momentum}
Sander, M.~E.; Ablin, P.; Blondel, M.; and Peyr{\'e}, G. 2021.
\newblock Momentum residual neural networks.
\newblock In \emph{International Conference on Machine Learning}, 9276--9287. PMLR.

\bibitem[{Suetake et~al.(2023)Suetake, Ikegawa, Saiin, and Sawada}]{suetake2023s3nn}
Suetake, K.; Ikegawa, S.-i.; Saiin, R.; and Sawada, Y. 2023.
\newblock S3NN: Time step reduction of spiking surrogate gradients for training energy efficient single-step spiking neural networks.
\newblock \emph{Neural Networks}, 159: 208--219.

\bibitem[{Tan, Wu, and Lu(2021)}]{tan2021improved}
Tan, P.-Y.; Wu, C.-W.; and Lu, J.-M. 2021.
\newblock An improved STBP for training high-accuracy and low-spike-count spiking neural networks.
\newblock In \emph{2021 Design, Automation \& Test in Europe Conference \& Exhibition (DATE)}, 575--580. IEEE.

\bibitem[{Wang et~al.(2023)Wang, Song, Wang, Xiao, Yang, Mei, and Zhang}]{wang2023ssf}
Wang, J.; Song, Z.; Wang, Y.; Xiao, J.; Yang, Y.; Mei, S.; and Zhang, Z. 2023.
\newblock Ssf: Accelerating training of spiking neural networks with stabilized spiking flow.
\newblock In \emph{Proceedings of the IEEE/CVF International Conference on Computer Vision}, 5982--5991.

\bibitem[{Wu et~al.(2018)Wu, Deng, Li, Zhu, and Shi}]{wu2018spatio}
Wu, Y.; Deng, L.; Li, G.; Zhu, J.; and Shi, L. 2018.
\newblock Spatio-temporal backpropagation for training high-performance spiking neural networks.
\newblock \emph{Frontiers in neuroscience}, 12: 331.

\bibitem[{Xiao et~al.(2022)Xiao, Meng, Zhang, He, and Lin}]{xiao2022online}
Xiao, M.; Meng, Q.; Zhang, Z.; He, D.; and Lin, Z. 2022.
\newblock Online training through time for spiking neural networks.
\newblock \emph{Advances in neural information processing systems}, 35: 20717--20730.

\bibitem[{Xu, Liu, and Yang(2023)}]{xu2023ultra}
Xu, C.; Liu, Y.; and Yang, Y. 2023.
\newblock Ultra-low latency spiking neural networks with spatio-temporal compression and synaptic convolutional block.
\newblock \emph{Neurocomputing}, 550: 126485.

\bibitem[{Yang and Chen(2023)}]{yang2023effective}
Yang, S.; and Chen, B. 2023.
\newblock Effective surrogate gradient learning with high-order information bottleneck for spike-based machine intelligence.
\newblock \emph{IEEE transactions on neural networks and learning systems}.

\bibitem[{Zhang et~al.(2025)Zhang, Wang, Xiao, Wei, Shan, Zhang, and Yang}]{zhang2025memory}
Zhang, D.; Wang, S.; Xiao, Y.; Wei, W.; Shan, Y.; Zhang, M.; and Yang, Y. 2025.
\newblock Memory-Free and Parallel Computation for Quantized Spiking Neural Networks.
\newblock In \emph{ICASSP 2025-2025 IEEE International Conference on Acoustics, Speech and Signal Processing (ICASSP)}, 1--5. IEEE.

\bibitem[{Zhang and Zhang(2024)}]{zhang2024memory}
Zhang, H.; and Zhang, Y. 2024.
\newblock Memory-efficient reversible spiking neural networks.
\newblock In \emph{Proceedings of the AAAI conference on artificial intelligence}, volume~38, 16759--16767.

\bibitem[{Zhang and Li(2019)}]{zhang2019spike}
Zhang, W.; and Li, P. 2019.
\newblock Spike-train level backpropagation for training deep recurrent spiking neural networks.
\newblock \emph{Advances in neural information processing systems}, 32.

\bibitem[{Zhao et~al.(2024)Zhao, Liu, Mangalam, Qian, Zohra, Alghannam, Malik, and Ghanem}]{zhao2024dr2net}
Zhao, C.; Liu, S.; Mangalam, K.; Qian, G.; Zohra, F.; Alghannam, A.; Malik, J.; and Ghanem, B. 2024.
\newblock Dr2Net: Dynamic reversible dual-residual networks for memory-efficient finetuning.
\newblock In \emph{Proceedings of the IEEE/CVF Conference on Computer Vision and Pattern Recognition}, 15835--15844.

\bibitem[{Zheng et~al.(2021)Zheng, Wu, Deng, Hu, and Li}]{zheng2021going}
Zheng, H.; Wu, Y.; Deng, L.; Hu, Y.; and Li, G. 2021.
\newblock Going deeper with directly-trained larger spiking neural networks.
\newblock In \emph{Proceedings of the AAAI conference on artificial intelligence}, volume~35, 11062--11070.

\end{thebibliography}


\end{document}